\documentclass[10pt, a4paper]{article}

\usepackage{lrec-coling2024}

\usepackage{tempora}
\usepackage{listings}

\usepackage{natbib}
\usepackage{multibib}
\makeatletter
\def\@mb@citenamelist{cite,citep,citet,citealp,citealt,citepalias,citetalias}
\makeatother
\newcites{languageresource}{~}

\usepackage{graphicx}
\usepackage{tabularx}
\usepackage{soul}
\usepackage{titlesec}
\titleformat{\section}{\normalfont\large\bfseries\center}{\thesection.}{1em}{}
\titleformat{\subsection}{\normalfont\SmallTitleFont\bfseries\raggedright}{\thesubsection.}{1em}{}
\titleformat{\subsubsection}{\normalfont\normalsize\bfseries\raggedright}{\thesubsubsection.}{1em}{}
\renewcommand\thesection{\arabic{section}}
\renewcommand\thesubsection{\thesection.\arabic{subsection}}
\renewcommand\thesubsubsection{\thesubsection.\arabic{subsubsection}}

\usepackage[table,xcdraw]{xcolor}
\usepackage{hyperref}
\definecolor{darkblue}{rgb}{0, 0, 0.5}
\hypersetup{colorlinks=true, citecolor=darkblue, linkcolor=darkblue, urlcolor=darkblue}

\usepackage{xstring}

\usepackage{color}

\usepackage{multirow,booktabs,url}
\usepackage[T2A,T1]{fontenc}
\usepackage[english, russian]{babel}
\usepackage{extdash}
\usepackage{xstring}
\usepackage{tipa}
\usepackage{tabularx,booktabs}

\addto\captionsrussian{
\renewcommand{\abstractname}{Abstract}

}

\newcommand{\scell}[2][c]{
\begin{tabular}[#1]{@{}c@{}}#2\end{tabular}}

\title{KazSAnDRA: \\Kazakh Sentiment Analysis Dataset of Reviews and Attitudes}

\name{Rustem Yeshpanov, Huseyin Atakan Varol}

\address{Institute of Smart Systems and Artificial Intelligence\\Nazarbayev University, Astana, Kazakhstan \\
\{rustem.yeshpanov, ahvarol\}@nu.edu.kz\\}

\abstract{
This paper presents KazSAnDRA, a dataset developed for Kazakh sentiment analysis that is the first and largest publicly available dataset of its kind. 
KazSAnDRA comprises an extensive collection of 180,064 reviews obtained from various sources and includes numerical ratings ranging from 1 to 5, providing a quantitative representation of customer attitudes. The study also pursued the automation of Kazakh sentiment classification through the development and evaluation of four machine learning models trained for both polarity classification and score classification. Experimental analysis included evaluation of the results considering both balanced and imbalanced scenarios. The most successful model attained an F$_1$-score of 0.81 for polarity classification and 0.39 for score classification on the test sets. The dataset and fine-tuned models are open access and available for download under the Creative Commons Attribution 4.0 International License (CC BY 4.0) through our GitHub repository.\\\newline
}

\begin{document}

\maketitleabstract

\section{Introduction}

In natural language processing, sentiment analysis is a widely employed text classification task that involves extracting the sentiment expressed by individuals towards a variety of entities that include products, services, organisations, individuals, issues, events, and topics together with their respective attributes~\citep{Liu2012}. In this context, sentiment represents the positive, negative, or neutral attitude of individuals conveyed through the extracted textual content~\citep{jurafsky2009}. 
Sentiment analysis demonstrates broad applicability across various domains, including marketing~\citep{fang2015sentiment}, social media~\citep{go2009twitter}, healthcare~\citep{Greaves251}, finance~\citep{abraham2018cryptocurrency}, and politics~\citep{abercrombie-batista-navarro-2020-parlvote}, among others.

Although research efforts in sentiment analysis for lower-resourced languages are gradually gaining momentum~\citep{mamta,tuan,gangula}, the English language continues to dominate as the primary focus of current research in this area~\citep{https://doi.org/10.1002/widm.1253}. 
This preference can be attributed to the abundant availability of linguistic resources, such as lexica, corpora, and dictionaries specifically tailored to English~\citep{MEDHAT20141093}.

With respect to Kazakh, an agglutinative Turkic language generally considered lower-resourced, research in the field of sentiment analysis has only recently come to the fore~\citep{sakenovich2016one}. Despite its importance, the literature dealing with sentiment analysis in Kazakh remains limited and includes only a few academic papers published within eight years. 
Furthermore, there is a complete absence of publicly accessible Kazakh sentiment analysis datasets, whether small or large, further underscoring the challenges in this field.

Our study aims to address the existing gaps in this field and contribute to its further advancement. 
Specifically, we present a dataset consisting of customer reviews in Kazakh, accompanied by corresponding attitude scores. 
The dataset comprises a total of 180,064 reviews collected from four domains. 

In the context of Kazakhstan, it is crucial to acknowledge the prevalent practice of code-switching between the Kazakh and Russian languages, as well as the ongoing shift from the Cyrillic to the Latin script. 
Consequently, Kazakh reviews may exhibit a combination of Cyrillic and Latin characters, incorporate a mixture of Russian and Kazakh vocabulary, or be solely recorded in the Cyrillic script with Russian characters substituting Kazakh ones. 
The dataset we present includes reviews containing both exclusive Kazakh vocabulary and words from other languages (Russian, English, and Arabic), making it the largest dataset available for Kazakh sentiment analysis.

We also developed and evaluated four machine learning models to automate the classification of Kazakh sentiments. The highest F$_1$-score on the test sets was 0.81 for polarity classification and 0.39 for score classification. 

The subsequent sections of this paper are structured as follows: \hyperref[sec2]{Section~2} presents a review of existing research in Kazakh sentiment analysis. \hyperref[sec3]{Section~3} is devoted to the detailing the process of developing the dataset. \hyperref[sec4]{Section~4} delves into the aspects of data pre-processing and partitioning, the score resampling techniques, the sentiment classification tasks, the models employed, the experimental design, and the metrics used for evaluation, and the corresponding results. \hyperref[sec5]{Section~5} focuses on a thorough discussion of the results. \hyperref[sec6]{Section~6} provides a conclusive summary and final remarks for the paper.

\section{Related Work}\label{sec2}

In recent years, remarkable progress has been made in addressing the limited resources available for the Kazakh language. Mussakhojayeva et al. have made significant contributions to this endeavour by presenting a text-to-speech synthesis corpus comprising a substantial 271 hours of speech data~\citep{mussakhojayeva-etal-2022-kazakhtts2}, as well as introducing the first industrial-scale open-source Kazakh speech corpus for automatic speech recognition~\citep{mussakhojayeva22_interspeech}. 
The latter corpus consists of 1,128 hours of transcribed audio data, comprising over 520 thousand utterances.
In a separate study,~\citet{yeshpanov-etal-2022-kaznerd} made a noteworthy contribution to the field of Kazakh natural language processing by introducing the largest dataset for Kazakh named entity recognition. This dataset comprises 112,702 sentences and 136,333 annotations for 25 distinct entity classes. 
In addition,~\citet{TOIGANBAYEVA2022116827} proposed an extensive dataset specifically tailored for handwritten text recognition in Kazakh. This dataset includes 3,000 handwritten exam papers, with 140,335 segmented images and 922,010 symbols. 

While the collective contributions to various speech and language processing tasks for Kazakh have undeniably enriched the available resources, thus creating new opportunities for research and development, progress in Kazakh sentiment analysis research has been comparatively slower. 
This discrepancy can be attributed to the limited availability of dedicated resources in this area.

In the earliest work found on Kazakh sentiment analysis~\citep{sakenovich2016one}, the researchers curated a dataset of 30,000 Russian news articles that were manually labelled. In addition, they labelled 10,000 Kazakh news articles, of which 3,021 were positive, 2,548 negative, and 4,431 neutral, to train a sentiment classifier. While the performance of the classifier for Russian was 86.3\%, it yielded relatively lower results for Kazakh, with an accuracy of 72.8\%. This result could possibly be attributed to the limited size of the training dataset and the absence of lemmatisation, which may have affected the overall performance.

In~\citet{Abdullin2017DEEPLM}, the research aimed to analyse opinions in short texts written in several languages, with a specific focus on English, Russian, and Kazakh~\citep{go2009twitter,rubtsova2014approach}. The authors presented an approach that utilised a deep recurrent neural network and bilingual word embeddings to effectively capture semantic features between words across these languages. By conducting sentiment analysis experiments on language pairs such as English-Russian and Russian-Kazakh, the authors achieved noteworthy performance, with a competitive accuracy rate of 72\% for Russian and a comparatively lower accuracy of 58\% for Kazakh. While the authors made the development codes available on their GitHub page for result reproducibility, it is worth noting that the repository does not include the Kazakh training data that were utilised.

In~\citet{8093531}, reviews of three hotels were collected from online travel platforms. The authors employed fuzzy logic~\citep{Zadeh1996FuzzyL} for sentiment analysis. However, it is important to acknowledge that the study does not provide precise information about the number of reviews collected and the accessibility of the data collected. In their later study~\citep{Yergesh2019SentimentAO}, the researchers presented an overview of the rule-based methods employed in sentiment analysis and the approaches utilised to determine the sentiment of Kazakh sentences by formalising morphological rules. To facilitate the determination of text polarity, a dictionary of approximately 11,000 emotional Kazakh words and phrases was manually compiled and annotated on a five-point scale [-2, 2]. In addition, semantic hypergraphs were used to describe ontological models of the morphological rules of the Kazakh language. As a result of this research, a morphological analyser was developed, enabling the extraction of morphological information from texts. The authors reported that they achieved a result of 83\% using their rule-based method, which they describe as ``good'', although it is interesting to note that this result was peculiarly compared to studies dedicated to sentiment analysis in Russian.

\citet{10.1007/978-3-030-24289-3_53} developed a method for analysing Kazakh texts related to terrorist threats. The research involved selecting social networks along with specific foreign Internet resources that disseminate terrorist content. Through this selection, a database was developed that comprised 1,200 entries. This database facilitated the detection of more than 50 similar sites. It is important to note, however, that access to the database is not possible. 

In~\citet{mutanov}, a dataset of news posts from the Kazakhstani media space was collected, comprising texts labelled across three sentiment classes: positive, negative, and neutral. The dataset includes 80,873 sentiment-labelled texts in Russian and 15,933 in Kazakh, revised by graduate students specialising in political science. While the paper describes in detail the steps of data pre-processing and the classification methods employed, it does not delve into the approach taken for news posts exhibiting multiple polarities (e.g., ``Controversial Policy Sparks Outrage and Support Among Citizens''), nor does it shed light on the provision of classification guidelines or the inter-annotator agreement. Additionally, while detailed classification metrics and confusion matrices are presented for Russian texts, analysis for Kazakh texts is notably absent.

In~\citet{Gimadi2021WebsentimentAO}, the aim of the study was to collect a dataset of 3,000 Kazakh reviews from the Google Play Store. However, the rationale behind the researcher's decision to manually label each of the collected reviews and subsequently compare the assigned scores with the original scores remains unclear.

More recently,~\citet{Rakhymzhanov2022ANAT} attempted to develop a Kazakh slang dictionary using a website\footnote{\url{https://janasozdik.kz}} and its associated Instagram page. The researcher utilised BeautifulSoup\footnote{\url{https://www.crummy.com/software/BeautifulSoup}} to collect slang word information, but due to the recent creation of the website, data availability was limited. This study breaks new ground in Kazakh sentiment analysis, and while the referenced slang dictionary provides amusing expressions, its practical usefulness as a reliable reference source can be questioned due to their infrequent use among native Kazakh speakers.

\citet{nurlybayeva2022kazakh} presented the construction of a bag-of-words model~\citep{Zhang2010} for sentiment classification of restaurant reviews into positive or negative categories. The researchers collected a dataset of 2,000 restaurant reviews from the 2GIS application website\footnote{\url{https://www.2gis.kz}} for analysis and model development. However, it should be noted that the dataset used in this study is not publicly accessible.

The last study on Kazakh sentiment analysis that we discuss in this paper is by~\citet{9945811}, who applied pre-trained BERT~\citep{bert} models, originally developed for multilingual and Turkish sentiment analysis, to Kazakh due to the lack of large labelled datasets in this language. The Kazakh dataset was collected from Facebook groups, a consumer complaints website\footnote{\url{https://zhalobikz.com}}, and the 2GIS website, with the reviews manually labelled and transliterated. The training dataset was very small (only 30 samples), but the experimental results showed that the multilingual BERT model outperformed the Turkish BERT model.

From the literature reviewed, it follows that while substantial efforts have been made in the fields of automatic speech recognition, text-to-speech synthesis, image-to-text conversion, and named entity recognition for Kazakh, resulting in extensive, high-quality and publicly available datasets, the same level of generosity seems to be lacking when it comes to research specifically focused on Kazakh sentiment analysis. In other words, the availability of sentiment analysis datasets for Kazakh is currently non-existent or significantly limited.

\section{Dataset Development}\label{sec3}

\subsection{Domains}
The source data for our dataset came from four domains: (1) digital mapping and navigation services (hereafter Mapping), (2) online marketplaces (hereafter Market), (3) an online library that serves as a source of books and audiobooks in Kazakh (hereafter Bookstore), and (4) an online store for Android devices that offers a diverse range of applications (hereafter Appstore).

\subsection{Data Collection}
The dataset was collected over a one-year period, from September 2022 to September 2023. Reviews from Mapping and Market were collected through manual means, while a \textit{BeautifulSoup} script was employed for the acquisition of reviews from Bookstore. The use of the Python package \textit{google-play-scraper}\footnote{\url{https://pypi.org/project/google-play-scraper}} facilitated the collection of reviews from Appstore.

All reviews were manually checked by a group of native Kazakh speakers. During the review process, it came to light that reviews contained recurrent instances of inappropriate content. However, in order to preserve the authenticity and integrity of the reviews, no alterations or removals were made to this content.

As a result, 8,897 reviews were obtained from Mapping, covering 407 institutions.  Market provided a considerable portion of 30,289 reviews, encompassing 8,418 unique items. Bookstore provided 5,805 reviews, comprising 3,792 audiobook reviews and 2,013 book reviews, resulting in a total of 1,026 unique audiobooks and books. Finally, Appstore provided 135,073 unique reviews of 1,759 Android applications and games. Of the users contributing to these reviews, 47,887 had a unique username, while the remaining 31,490 users remained anonymous. 

Each review was accompanied by a numerical rating from 1 to 5, providing a measurable representation of individuals' attitudes. Consequently, we named the dataset KazSAnDRA~/kә\textprimstress s\ae ndrә/, an acronym for the \textbf{Kaz}akh \textbf{S}entiment \textbf{An}alysis \textbf{D}ataset of \textbf{R}eviews and \textbf{A}ttitudes, reflecting its purpose and content. The total number of reviews collected was 180,064. Table~\ref{tab:source_stats} provides information about the distribution of reviews across different scores and domains.

\begin{table}[!ht]
\fontsize{8.5}{10.2}\selectfont
\setlength\tabcolsep{0.11cm}
\begin{tabular}{l|rrrrr|r}
\toprule
\multicolumn{1}{c|}{\multirow{2}{*}{\textbf{Domain}}} & \multicolumn{5}{c|}{\textbf{Score}} & \multicolumn{1}{c}{\multirow{2}{*}{\textbf{Total}}}\\
\multicolumn{1}{c|}{} & \multicolumn{1}{c}{\textbf{1}} & \multicolumn{1}{c}{\textbf{2}} & \multicolumn{1}{c}{\textbf{3}} & \multicolumn{1}{c}{\textbf{4}} & \multicolumn{1}{c|}{\textbf{5}} & \multicolumn{1}{c}{}\\
\midrule
Appstore & 22,547 & 4,202 & 5,758 & 7,949 & 94,617 & \multicolumn{1}{r}{{\textbf{135,073}}} \\
Bookstore & 686 & 107 & 222 & 368 & 4,422 & \multicolumn{1}{r}{{\textbf{5,805}}} \\
Mapping & 959 & 270 & 369 & 525 & 6,774 & \multicolumn{1}{r}{{\textbf{8,897}}} \\
Market & 1,043 & 350 & 913 & 2,775 & 25,208 & \multicolumn{1}{r}{{\textbf{30,289}}} \\
\midrule
\multicolumn{1}{c|}{{\textbf{Total}}} & \textbf{25,235} & \textbf{4,929} & \textbf{7,262} & \textbf{11,617}  & \textbf{131,021} & \multicolumn{1}{r}{{\textbf{180,064}}} \\
\bottomrule
\end{tabular}
\caption{Domain and score statistics\label{tab:source_stats}}
\end{table}

\subsection{Variations of Kazakh Reviews}
In Kazakhstan, code-switching between the Kazakh and Russian languages has been observed~\citep{Pavlenko2008}, alongside an ongoing shift from the Cyrillic to the Latin script. Consequently, reviews regarded as Kazakh can take different forms: (a) solely Kazakh words written in the Kazakh Cyrillic script, (b) Kazakh words written in Latin script, (c) a combination of Cyrillic and Latin characters, (d) a mixture of Russian and Kazakh words, or (e) a text entirely in the Cyrillic script with Russian characters replacing Kazakh characters, among other possible variants. Table~\ref{tab:manifestations} provides examples of actual reviews, demonstrating their appropriate representation in accordance with Kazakh spelling rules and the use of the Kazakh Cyrillic script, accompanied by their correct form in English. Table~\ref{tab:cyr_lat} shows the number of reviews with percentage of Cyrillic and Latin characters per review.

\begin{table}[!ht]
\setlength\tabcolsep{0.14cm}
\centering
\begin{tabular}{c|rrrr}
\toprule
\textbf{Character} & \textbf{0–25\%} & \textbf{26–50\%} & \textbf{51–75\%} & \textbf{76–100\%} \\
\midrule
Cyrillic & 67 & 399 & 1,694 & 170,233 \\
Latin & 5,374 & 1,114 & 246 & 2,617 \\
\bottomrule
\end{tabular}
\caption{Review counts by Cyrillic and Latin character percentages\label{tab:cyr_lat}}
\end{table}

\begin{table*}[!t]
\centering
\setlength{\tabcolsep}{3.8mm}
\begin{center}
\begin{tabular}{clll}
\toprule
\multicolumn{1}{l}{} & \multicolumn{1}{l}{\textbf{Actual review}} & \textbf{Correct form (Kazakh)} & \textbf{Correct form (English)} \\ 
\midrule
a & \textit{керемет кітап} &  \textit{керемет кітап} & \textit{a wonderful book} \\
b & \textit{keremet} & \textit{керемет} & \textit{wonderful} \\
c & \textit{jok кітап} & \textit{кітап жоқ} & \textit{no books} \\
d & \textit{Осы приложениеге көп рахмет!} & \textit{Осы қолданбаға көп рақмет!} & \textit{Many thanks to this app!} \\
e & \textit{Кушти!} & \textit{Күшті!} & \textit{Great!} \\
\bottomrule
\end{tabular}
\end{center}
\caption{Kazakh review variations\label{tab:manifestations}}
\end{table*}

\subsection{Sentiment Classification Tasks}
To evaluate the effectiveness of KazSAnDRA, we utilised the dataset for two tasks: (a) polarity classification (PC), which involves predicting whether a review is positive or negative, and (b) score classification (SC), which involves predicting the score of a review on a scale of 1 to 5. In the PC task only, reviews with an original score of 1 or 2 were designated as negative and subsequently assigned a new score of 0. In contrast, reviews with an original score of 4 and 5 were classified as positive and assigned a new score of 1. Reviews with an original score of 3 were categorised as neutral and excluded from the task.

\subsection{Data Pre-Processing}
Irrespective of the task for which the dataset was intended, the data pre-processing stage involved several essential steps aimed at preserving the integrity and uniformity of the dataset. First, all emojis were systematically removed from the text to eliminate potential noise. Subsequently, to ensure consistency and ease of analysis, all reviews were lowercased. The elimination of punctuation helped to streamline the text for further processing. In addition, the characters for line break (\textbackslash n), tab (\textbackslash t), and carriage return (\textbackslash r)  were removed to avoid interference with subsequent computations. To enhance readability and minimise unnecessary mismatches, multiple spaces were uniformly replaced with a single space.
Furthermore, it is important to note that in the Kazakh language, consecutive characters are allowed to occur in pairs (e.g., \textit{айттым} ``I said'', \textit{құжаттар} ``documents'', \textit{қосса} ``if s/he adds'') but not in larger clusters. Hence, to conform to this linguistic feature, any consecutive characters that appeared repeatedly were reduced to two instances (e.g., \textit{кееррреемееетт} to \textit{кеерреемеетт}). Lastly, all duplicate entries, defined as reviews sharing identical text and scores, were removed. The total numbers of reviews following pre-processing were 167,961 for PC and 175,158 for SC.

\subsection{Data Partitioning}
To ensure consistency and reproducibility of our experimental results across different research groups, KaZSAnDRA was divided into training (Train), validation (Valid), and test (Test) sets, maintaining a ratio of 80/10/10. 
The division ensured a balanced and proportional distribution of review scores and domains across the sets.

Tables~\ref{tab:set_sources_pc} and~\ref{tab:set_sources_sc} present the distribution of reviews across the three sets based on the domains for the PC and SC tasks, respectively, after pre-processing, including the total number of reviews as well as the respective counts for each set pertaining to both tasks. Tables~\ref{tab:set_scores_pc} and~\ref{tab:set_scores_sc} present the distribution of reviews across the three sets in terms of their scores for the PC and SC tasks, in turn.

\begin{figure}[!ht]
\begin{center}
\includegraphics[width=1\linewidth,trim={0.7cm 0.7cm 0.7cm 0.7cm},clip=true]{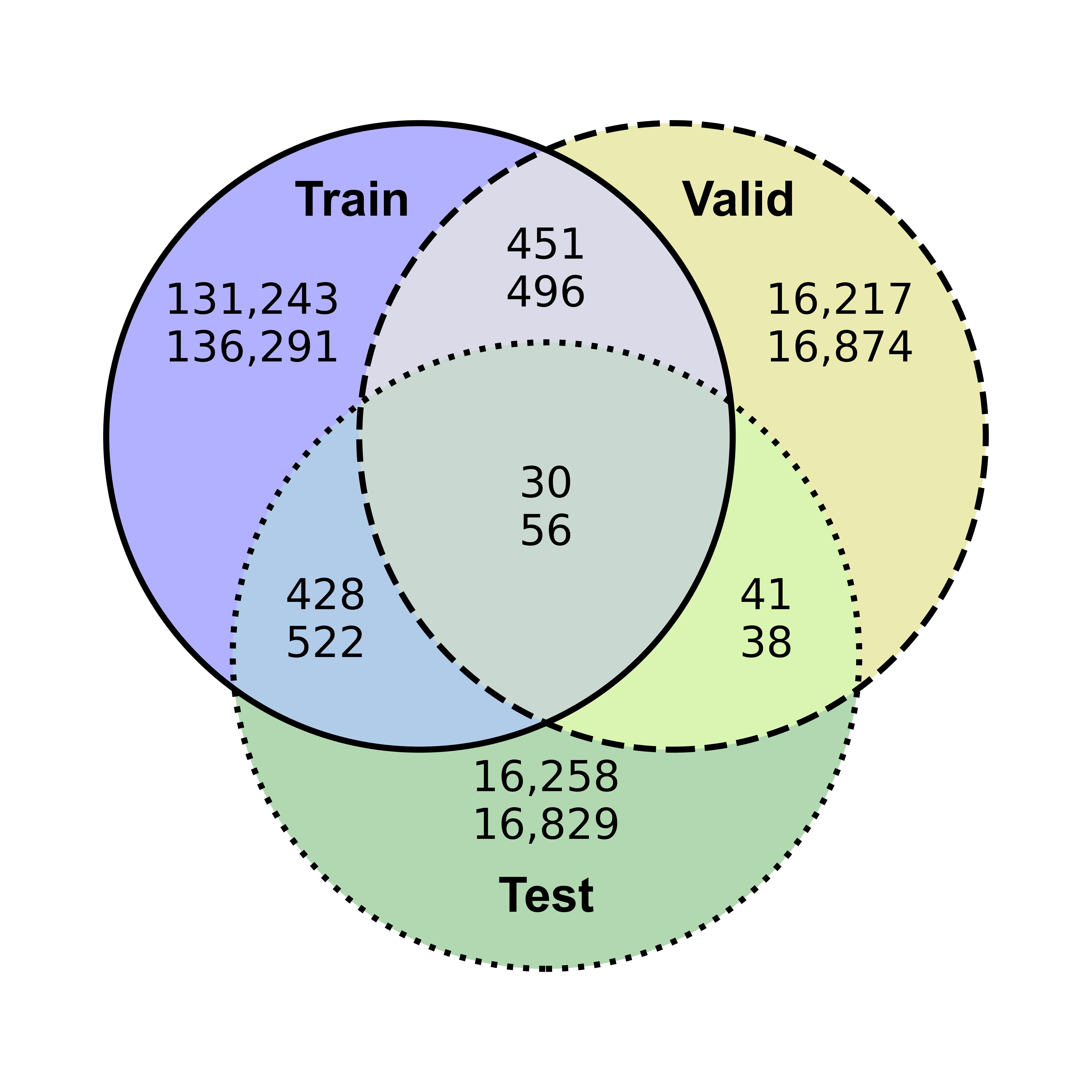}
\vspace{-1cm}
\caption{Unique reviews across the sets for  PC (top) and SC (bottom)}
\label{fig:venn_pc_and_sc}
\end{center}
\end{figure}

Furthermore, an analysis of the KazSAnDRA dataset was conducted to extract unique reviews (i.e., reviews with distinct textual content) per classification task. The numbers of unique reviews in the training, validation, and test sets for PC were 132,152, 16,739, and 16,757, respectively. For SC, the corresponding counts were 137,365, 17,464, and 17,445. 
The intersection between the training, validation, and test sets was then computed, as depicted in Figure~\ref{fig:venn_pc_and_sc}. 
Significantly, over 96\% of the unique reviews present in the test sets for both PC and SC tasks did not occur in either the training or validation sets. This substantial discrepancy corroborated the appropriateness of utilising the test sets to effectively evaluate the generalisation capabilities of sentiment classification models.

\begin{table}[!ht]
\small
\setlength\tabcolsep{0.1cm}
\begin{center}
\begin{tabularx}{\columnwidth}{c|rr|rr|rr}
\toprule
\multirow{2}{*}{\textbf{Domain}} & \multicolumn{2}{c|}{\textbf{Train}}& \multicolumn{2}{c|}{\textbf{Valid}}& \multicolumn{2}{c}{\textbf{Test}} \\
&\multicolumn{1}{c}{\textbf{\#}}&\multicolumn{1}{c|}{\textbf{\%}}&\multicolumn{1}{c}{\textbf{\#}}&\multicolumn{1}{c|}{\textbf{\%}}&\multicolumn{1}{c}{\textbf{\#}}&\multicolumn{1}{c}{\textbf{\%}}\\
\midrule
Appstore & 101,477 & 75.52 & 12,685 & 75.52 & 12,685 & 75.52 \\
Market & 22,561 & 16.79 & 2,820 & 16.79 & 2,820 & 16.79 \\
Mapping & 6,509 & 4.84 & 813 & 4.84 & 814 & 4.85 \\
Bookstore & 3,821 & 2.84 & 478 & 2.85 & 478 & 2.85 \\
\midrule
\textbf{Total} & \textbf{134,368} & \textbf{100} & \textbf{16,796} & \textbf{100} & \textbf{16,797} & \textbf{100} \\
\bottomrule
\end{tabularx}
\caption{Domains across the sets for PC \label{tab:set_sources_pc}}
\end{center}
\end{table}

\begin{table}[!ht]
\small
\setlength\tabcolsep{0.1cm}
\begin{center}
\begin{tabularx}{\columnwidth}{c|rr|rr|rr}
\toprule
\multirow{2}{*}{\textbf{Domain}} & \multicolumn{2}{c|}{\textbf{Train}}& \multicolumn{2}{c|}{\textbf{Valid}}& \multicolumn{2}{c}{\textbf{Test}} \\
&\multicolumn{1}{c}{\textbf{\#}}&\multicolumn{1}{c|}{\textbf{\%}}&\multicolumn{1}{c}{\textbf{\#}}&\multicolumn{1}{c|}{\textbf{\%}}&\multicolumn{1}{c}{\textbf{\#}}&\multicolumn{1}{c}{\textbf{\%}}\\
\midrule
Appstore & 106,058 & 75.69 & 13,258 & 75.69 & 13,257 & 75.69 \\[0.025cm]
Market & 23,278 & 16.61 & 2,909 & 16.61 & 2,910 & 16.61 \\[0.025cm]
Mapping & 6,794 & 4.85 & 849 & 4.85 & 849 & 4.85 \\[0.025cm]
Bookstore & 3,996 & 2.85 & 500 & 2.85 & 500 & 2.85 \\
\midrule
\textbf{Total} & \textbf{140,126} & \textbf{100} & \textbf{17,516} & \textbf{100} & \textbf{17,516} & \textbf{100} \\
\bottomrule
\end{tabularx}
\caption{Domains across the sets for SC \label{tab:set_sources_sc}}
\end{center}
\end{table}

\begin{table}[!ht]
\small
\setlength\tabcolsep{0.13cm}
\begin{center}
\begin{tabularx}{\columnwidth}{c|rr|rr|rr}
\toprule
\multirow{2}{*}{\textbf{\scell{Score}}} & \multicolumn{2}{c|}{\textbf{Train}} & \multicolumn{2}{c|}{\textbf{Valid}} & \multicolumn{2}{c}{\textbf{Test}} \\
& \multicolumn{1}{c}{\textbf{\#}} & \multicolumn{1}{c|}{\textbf{\%}} & \multicolumn{1}{c}{\textbf{\#}} & \multicolumn{1}{c|}{\textbf{\%}} & \multicolumn{1}{c}{\textbf{\#}} & \multicolumn{1}{c}{\textbf{\%}} \\ 
\midrule
1 & 110,417 & 82.18 & 13,801 & 82.17 & 13,804 & 82.18 \\
0 & 23,951 & 17.82 & 2,995 & 17.83 & 2,993 & 17.82 \\
\midrule
\textbf{Total} & \textbf{134,368} & \textbf{100} & \textbf{16,796} & \textbf{100} & \textbf{16,797} & \textbf{100} \\
\bottomrule
\end{tabularx}
\caption{Scores across the sets for PC\label{tab:set_scores_pc}}
\end{center}
\end{table}

\begin{table}[!ht]
\small
\setlength\tabcolsep{0.13cm}
\begin{center}
\begin{tabularx}{\columnwidth}{c|rr|rr|rr}
\toprule
\multirow{2}{*}{\textbf{\scell{Score}}} & \multicolumn{2}{c|}{\textbf{Train}} & \multicolumn{2}{c|}{\textbf{Valid}} & \multicolumn{2}{c}{\textbf{Test}} \\
& \multicolumn{1}{c}{\textbf{\#}} & \multicolumn{1}{c|}{\textbf{\%}} & \multicolumn{1}{c}{\textbf{\#}} & \multicolumn{1}{c|}{\textbf{\%}} & \multicolumn{1}{c}{\textbf{\#}} & \multicolumn{1}{c}{\textbf{\%}} \\ 
\midrule
5 & 101,302 & 72.29 & 12,663 & 72.29 & 12,663 & 72.29 \\
1 & 20,031 & 14.29 & 2,504 & 14.30 & 2,504 & 14.30 \\
4 & 9,115 & 6.50 & 1,140 & 6.51 & 1,139 & 6.50 \\
3 & 5,758 & 4.11 & 719 & 4.10 & 720 & 4.11 \\
2 & 3,920 & 2.80 & 490 & 2.80 & 490 & 2.80 \\
\midrule
\textbf{Total} & \textbf{140,126} & \textbf{100} & \textbf{17,516} & \textbf{100} & \textbf{17,516} & \textbf{100} \\
\bottomrule
\end{tabularx}
\caption{Scores across the sets for SC\label{tab:set_scores_sc}}
\end{center}
\end{table}

\subsection{Score Resampling}

Table~\ref{tab:source_stats} shows the score distribution in KazSAnDRA, indicating a significant imbalance. 
This raises concerns about biassed model performance, favouring the majority scores and neglecting underrepresented scores. Therefore, our study included analysis of results obtained from both balanced and imbalanced data, which had previously been employed with varying degrees of success (see~\citealp{burns2011sentiment,mutanov}).

In response to the data imbalance in our training data, we employed random oversampling (ROS) and random undersampling (RUS) to balance the representation of classes by creating new samples for the smaller class to align with the majority class count and eliminating samples from the larger class to match the minority class count, respectively~\citep{Ramentol2012}. In this study, we deferred the investigation of alternative approaches (e.g., data augmentation through back-translation) for future research. The resulting training sets for both tasks are detailed in Tables~\ref{tab:score_resampling_pc} and ~\ref{tab:score_resampling_sc}. 

\begin{table}[!ht]
\small
\center
\setlength\tabcolsep{0.3cm}
\begin{tabular}{c|cc|r}
\toprule
\multirow{2}{*}{\textbf{Score}} & \multicolumn{2}{c|}{\textbf{Balanced}} & \multirow{2}{*}{\textbf{Imbalanced}} \\
& \textbf{ROS} & \textbf{RUS} &  \\
\midrule
0 & 110,417 & 23,951 & 23,951 \\
1 & 110,417 & 23,951 & 110,417 \\
\bottomrule
\end{tabular}
\caption{Training sets for PC}
\label{tab:score_resampling_pc}
\end{table}

\begin{table}[!ht]
\small
\center
\setlength\tabcolsep{0.3cm}
\begin{tabular}{c|cc|r}
\toprule
\multirow{2}{*}{\textbf{Score}} & \multicolumn{2}{c|}{\textbf{Balanced}} & \multirow{2}{*}{\textbf{Imbalanced}} \\
& \textbf{ROS} & \textbf{RUS} &  \\
\midrule
1 & 101,302 & 3,920 & 20,031 \\
2 & 101,302 & 3,920 & 3,920 \\
3 & 101,302 & 3,920 & 5,758 \\
4 & 101,302 & 3,920 & 9,115 \\
5 & 101,302 & 3,920 & 101,302 \\
\bottomrule
\end{tabular}
\caption{Training sets for SC}
\label{tab:score_resampling_sc}
\end{table}

\subsection{Dataset Organisation}
The dataset comprises ten CSV files. Files ``01'' to ``05'' are associated with PC, while files ``06'' to ``10'' are related to SC. Different training set variations are indicated by the suffixes ``ib'' for imbalanced data, ``ros'' for random oversampling, and ``rus'' for random undersampling. Each file includes records containing a custom review identifier (\texttt{custom\_id}), the original review text (\texttt{text}), the pre-processed review text (\texttt{text\_cleaned}), the corresponding review score (\texttt{label}), and the domain information (\texttt{domain}). The dataset can be conveniently downloaded from our GitHub repository.\footnote{\url{https://github.com/IS2AI/KazSAnDRA}\label{ft:github}}

\begin{table*}[!t]
\fontsize{7.5}{9.6}\selectfont
\setlength\tabcolsep{0.06cm}
\begin{tabular}{cccccccccccccccccccccccccccccc}
\toprule
\multirow{5}{*}{\textbf{Model}} & \multicolumn{14}{c}{\textbf{Polarity Classification}} &  & \multicolumn{14}{c}{\textbf{Score Classification}} \\
\cmidrule{2-15} \cmidrule{17-30}
& \multicolumn{4}{c}{\textbf{Balanced (ROS)}} &  & \multicolumn{4}{c}{\textbf{Balanced (RUS)}} &  & \multicolumn{4}{c}{\textbf{Imbalanced}} &  & \multicolumn{4}{c}{\textbf{Balanced (ROS)}} &  & \multicolumn{4}{c}{\textbf{Balanced (RUS)}} &  & \multicolumn{4}{c}{\textbf{Imbalanced}} \\
\cmidrule{2-5} \cmidrule{7-10} \cmidrule{12-15}
\cmidrule{17-20} \cmidrule{22-25} \cmidrule{27-30}
& \textbf{A} & \textbf{P} & \textbf{R} & \textbf{F$_1$} &  & \textbf{A} & \textbf{P} & \textbf{R} & \textbf{F$_1$} &  & \textbf{A} & \textbf{P} & \textbf{R} & \textbf{F$_1$} & & \textbf{A} & \textbf{P} & \textbf{R} & \textbf{F$_1$} &  & \textbf{A} & \textbf{P} & \textbf{R} & \textbf{F$_1$} &  & \textbf{A} & \textbf{P} & \textbf{R} & \textbf{F$_1$} \\
\cmidrule{2-5} \cmidrule{7-10} \cmidrule{12-15}
\cmidrule{17-20} \cmidrule{22-25} \cmidrule{27-30}
\textbf{mBERT} & 0.84 & 0.74 & 0.83 & 0.77 &  & 0.85 & 0.76 & 0.82 & 0.78 &  & 0.89 & 0.82 & 0.79 & 0.80 &  & 0.67 & 0.34 & 0.36 & 0.35 &  & 0.63 & 0.35 & 0.39 & 0.36 &  & 0.77 & 0.44 & 0.36 & 0.37 \\
\textbf{XLM-R} & 0.86 & 0.76 & 0.83 & 0.79 &  & 0.85 & 0.75 & 0.83 & 0.78 &  & \textbf{0.89} & \textbf{0.81} & \textbf{0.81} & \textbf{0.81} &  & 0.58 & 0.36 & 0.42 & 0.36 &  & 0.66 & 0.36 & 0.41 & 0.37 &  & \textbf{0.77} & \textbf{0.42} & \textbf{0.37} & \textbf{0.39} \\
\textbf{RemBERT} & 0.88 & 0.79 & 0.82 & 0.81 &  & 0.87 & 0.78 & 0.82 & 0.80 &  & \textbf{0.89} & \textbf{0.81} & \textbf{0.82} & \textbf{0.81} &  & 0.73 & 0.37 & 0.36 & 0.36 &  & 0.62 & 0.35 & 0.40 & 0.35 &  & \textbf{0.76} & \textbf{0.41} & \textbf{0.38} & \textbf{0.39} \\
\textbf{mBART-50} & 0.87 & 0.77 & 0.79 & 0.78 &  & 0.81 & 0.72 & 0.81 & 0.74 &  & 0.89 & 0.82 & 0.78 & 0.80 &  & 0.74 & 0.36 & 0.34 & 0.35 &  & 0.55 & 0.36 & 0.41 & 0.34 & & 0.77 & 0.42 & 0.37 & 0.38 \\
\bottomrule
\end{tabular}
\caption{PC and SC results on the test sets}
\label{tab:pc_and_sc_results}
\end{table*}

\section{Experiment}\label{sec4}

\subsection{Sentiment Classification Models}
For the evaluation of KazSAnDRA, we utilised four multilingual machine learning models, all incorporating the Kazakh language and accessible through the Hugging Face Transformers framework~\citep{wolf-etal-2020-transformers}. The framework was chosen for its state-of-the-art pre-trained models, user-friendly interface, and collaborative ecosystem. Details on the implementation of the sentiment classification model are available on our GitHub repository.\textsuperscript{\ref{ft:github}}

\textbf{mBERT}
is a case-insensitive variant of the multilingual BERT~\citep{bert} model. This model comprises about 168 million parameters and has been pre-trained on a corpus of 102 languages.

\textbf{XLM-R}
We leveraged the XLM-RoBERTa model~\citep{DBLP:conf/acl/ConneauKGCWGGOZ20}, a multilingual variant of RoBERTa~\citep{liu2019roberta}. The rationale for selecting this model stems from its substantial parameter count, exceeding that of BERT by over fivefold (561M), and its pre-training on the CommonCrawl corpus encompassing 100 languages.

\textbf{RemBERT}
The rebalanced multilingual BERT model~\citep{chung2021rethinking} is a multilingual encoder pre-trained on Wikipedia over 104 languages. RemBERT exhibits superior performance compared to the similarly sized XLM-R, despite being trained on 3.5 times fewer tokens. 

\textbf{mBART-50}
~\citep{tang2020multilingual} is a multilingual sequence-to-sequence model built on the foundation of the original mBART model~\citep{liu-etal-2020-multilingual-denoising}. This extended version was thoughtfully augmented with an additional 25 languages, bringing the total number of languages supported to 50.

\subsection{Experimental Setup}
All four models were fine-tuned using both the balanced and imbalanced training sets, while the hyperparameters were refined using the validation set. The final and most optimal models were evaluated on the test sets. The fine-tuning of the models was executed on a single A100 GPU hosted on an NVIDIA DGX A100 machine. The initial learning rate was set at $10^{-5}$; the weight decay rate was set at $10^{-3}$. Early stopping was employed, executed when the F$_1$-score exhibited no improvement for three consecutive epochs. We set the batch size to 32 (mBERT, XLM-R, RemBERT) or 16 (mBART-50) and applied 800 warm-up steps.

\subsection{Performance Metrics}
Several conventional metrics were used to evaluate the performance of the models, including accuracy (A), precision (P), recall (R), and F$_1$-score (F$_1$). 
Given the imbalanced nature of the dataset, where all classes carry equal importance, we opted for macro-averaging, calculated from the arithmetic (i.e., unweighted) mean of all F$_1$-scores per class, and thus ensuring equal treatment of all classes during the evaluation, resulting in a stronger penalty if the model performs worse on minority classes~\citep{jurafsky2009,yang2001study}.

\subsection{Experiment Results}

Table~\ref{tab:pc_and_sc_results} presents the performance of the four models on KazSAnDRA test sets for the PC and SC tasks. XLM-R and RemBERT consistently outperformed mBERT and mBART-50 across various training scenarios. The highest F$_1$-scores of 0.81 (PC) and 0.39 (SC) were achieved by both XLM-R and RemBERT in the imbalanced training scenario. In five out of six training scenarios, RemBERT achieved the highest F$_1$-scores, while XLM-R led in four. Table~\ref{tab:epochs} presents data on the number of epochs required to train models for the PC and SC tasks, considering both balanced (ROS, RUS) and imbalanced (IB) training data scenarios.
Tables~\ref{tab:rembert_cm_pc}–\ref{tab:rembert_domain} summarise the performance of RemBERT in the imbalanced training context, with a detailed analysis following in the subsequent section. 

\begin{table}[!ht]
\setlength\tabcolsep{0.15cm}
\begin{tabular}{cccccccc}
\toprule
\multirow{3}{*}{\textbf{Model}} & \multicolumn{3}{c}{\textbf{PC}} &  & \multicolumn{3}{c}{\textbf{SC}} \\
\cmidrule{2-4} \cmidrule{6-8}
& \textbf{ROS} & \textbf{RUS} & \textbf{IB} &  & \textbf{ROS} & \textbf{RUS} & \textbf{IB} \\
\cmidrule{2-4} \cmidrule{6-8}
\textbf{mBERT} & 4 & 7  & 6 &  & 8 & 10 & 11 \\
\textbf{XLM-R} & 5 & 7  & 5 &  & 4 & 9 & 16 \\
\textbf{RemBERT} & 4 & 5 & 5 &  & 6 & 6 & 9 \\
\textbf{mBART-50} & 5  & 7 & 5 &  & 8 & 7 & 5 \\
\bottomrule
\end{tabular}
\caption{Number of training epochs for models}
\label{tab:epochs}
\end{table}

\section{Discussion}\label{sec5}
Scores in the SC task were lower than in the PC task for all models, possibly due to its increased complexity involving five-way classification.
Table~\ref{tab:rembert_cm_pc} shows that, in the PC task, the RemBERT model had stronger performance in identifying positive reviews, but had a notable drawback misclassifying 1,036 positive reviews as negative, indicating a relatively high number of false negatives.

\begin{table}[!ht]
\setlength\tabcolsep{0.380cm}
\begin{tabular}{crrr}
\toprule
\multicolumn{4}{c}{\textbf{Polarity Classification}} \\
\midrule
\begin{tabular}[c]{@{}c@{}}\textbf{predicted} $\mathbf{\rightarrow}$\\ \textbf{actual} $\mathbf{\downarrow}$\end{tabular} & \multicolumn{1}{c}{\textbf{0}} & \multicolumn{1}{c}{\textbf{1}} & \textbf{Total} \\
\textbf{0} & 2,155 & 838 & 2,993 \\
\textbf{1}& 1,036 & 12,768 & 13,804 \\
\bottomrule
\end{tabular}
\caption{RemBERT PC results}
\label{tab:rembert_cm_pc}
\end{table}

In the SC task (Table~\ref{tab:rembert_cm_sc}), it appears that the model had higher accuracy in identifying reviews with scores at the extremes (1 and 5) compared to the middle scores (2, 3, and 4). The model had particularly low accuracy in identifying reviews with a score of 2, with only 55 true positives. The reason for this is most likely that the training data contained many more reviews with scores of 1 and 5 than the middle scores (see Table~\ref{tab:source_stats}).

\begin{table}[!h]
\setlength\tabcolsep{0.10cm}
\begin{tabular}{crrrrrr}
\toprule
\multicolumn{7}{c}{\textbf{Score Classification}} \\
\midrule
\begin{tabular}[c]{@{}c@{}}\textbf{predicted} $\mathbf{\rightarrow}$\\ \textbf{actual} $\mathbf{\downarrow}$\end{tabular} & \multicolumn{1}{c}{\textbf{1}} & \multicolumn{1}{c}{\textbf{2}} & \multicolumn{1}{c}{\textbf{3}} & \multicolumn{1}{c}{\textbf{4}} & \multicolumn{1}{c}{\textbf{5}} & \multicolumn{1}{c}{\textbf{Total}} \\
\textbf{1} & 1,379 & 145 & 132 & 64 & 784 & 2,504 \\
\textbf{2} & 182 & 55 & 56 & 25 & 172 & 490 \\
\textbf{3} & 173 & 54 & 118 & 65 & 310 & 720 \\
\textbf{4} & 110 & 39 & 90 & 169 & 731 & 1,139 \\
\textbf{5} & 564 & 59 & 165 & 297 & 11,578 & 12,663 \\
\bottomrule
\end{tabular}
\caption{RemBERT SC results}
\label{tab:rembert_cm_sc}
\end{table}

In addition, the model exhibited a tendency for misclassifying reviews with scores other than 5 as if they had a rating of 5. This also seems to be related to the preponderance of reviews with a score of 5, causing the model to have a bias towards this score. The substantial disparity between the number of positive and negative reviews can be attributed to the fact that the reviewed items predominantly represent highly popular or top-rated selections. It is therefore to be expected that such items naturally receive a significantly higher number of reviews than less popular or lower-rated items~\citep{aly-atiya-2013-labr}.

\begin{table}[!ht]
\fontsize{9.5}{11.4}\selectfont
\setlength\tabcolsep{0.075cm}
\begin{tabular}{cccccccccc}
\toprule
\multirow{3}{*}{\textbf{Domain}} & \multicolumn{4}{c}{\textbf{PC}} &  & \multicolumn{4}{c}{\textbf{SC}} \\
\cmidrule{2-5} \cmidrule{7-10}
& \textbf{A} & \textbf{P} & \textbf{R} & \textbf{F$_1$} &  & \textbf{A} & \textbf{P} & \textbf{R} & \textbf{F$_1$} \\
\cmidrule{2-5} \cmidrule{7-10}
\textbf{Appstore} & 0.87 & 0.80 & 0.81 & 0.80 &  & 0.74 & 0.41 & 0.37 & 0.38 \\
\textbf{Bookstore} & 0.86 & 0.75 & 0.80 & 0.77 &  & 0.73 & 0.34 & 0.32 & 0.32 \\
\textbf{Mapping} & 0.92 & 0.84 & 0.88 & 0.86 &  & 0.80 & 0.42 & 0.41 & 0.41 \\
\textbf{Market} & 0.97 & 0.84 & 0.91 & 0.87 &  & 0.82 & 0.43 & 0.41 & 0.42 \\
\bottomrule
\end{tabular}
\caption{RemBERT results by domain}
\label{tab:rembert_domain}
\end{table}

In Table~\ref{tab:rembert_domain}, an interesting observation is that the model exhibited more accurate classification in both tasks for reviews from the Mapping and Market domains, which were manually collected, as opposed to reviews from the other two domains acquired through automated means. This observation suggests that the moderators may have selectively collected reviews with higher readability, fewer spelling and grammar errors, and reduced instances of code-switching and inappropriate content during the review collection process. The availability of cleaner, less noisy data could have positively influenced the performance of the model in classifying Kazakh reviews.

It is also important to recognise that the poorer performance on the SC task may not have solely stemmed from the increased complexity and challenges inherent in multi-class classification tasks. It could also be an indicator of the quality of the reviews present within the dataset. Recall that the main objective of this study was to develop a dataset that includes a diverse array of Kazakh reviews of different products and services, which, in turn, would hopefully facilitate in-depth research in Kazakh sentiment analysis. Nevertheless, we frankly admit that certain aspects, such as the correction of spelling errors, the effective handling of frequent code-switching between Kazakh and Russian, and the application of lemmatisation techniques, were not explicitly addressed and may have resulted in the lower performance of the models. These specific challenges offer promising opportunities for future investigations to improve the quality and linguistic processing capabilities of the dataset.

Upon addressing the aforementioned aspects, data augmentation techniques, such as back-translation, could be considered as possible alternatives to ROS and RUS, which were used in our study to address the data imbalance issue. 
The experimental findings suggest that of the four models trained on data balanced using the mentioned techniques only RemBERT exhibited improvement, albeit solely in the PC task~\citep{burns2011sentiment}.

Another challenge that may have caused the low performance on the SC task lies in the pronounced dependence of classification on the discretion of the author of a review~\citep{smetanin2021deep}. The potential introduction of inaccuracies during the process of assigning ratings by the author could engender misclassification of the final labels within the dataset. For instance, an ostensibly positive review may paradoxically carry a score of 1; conversely, a review strongly critical of a product may be concomitantly associated with a high rating of 5. The absence of standardised criteria for sentiment labelling leads to a subjective, intuitive approach by individual authors and thus to considerable variability in the assignment of ratings between authors. This underscores the exigency of formulating sentiment annotation guidelines in and, more importantly, for the Kazakh language, which can serve as a framework for future research in this area.

\section{Conclusion}\label{sec6}
The aim of this study was to create an extensive dataset for Kazakh sentiment analysis. The result is KazSAnDRA, the first and largest publicly available dataset for Kazakh. Comprising 180,064 reviews from four domains, KazSAnDRA includes numerical ratings from 1 to 5 that quantitatively represent customers' attitudes. To automate Kazakh sentiment classification, we developed and evaluated four machine learning models for both polarity and score classification. The experimental analysis involved examining the results obtained with both balanced and imbalanced training data. The most successful model achieved an F$_1$-score of 0.81 for the polarity classification task and 0.39 for the score classification task on the test sets. In the future, we plan to improve KazSAnDRA by addressing spelling errors and effectively handling code-switching phenomena. These improvements will facilitate the use of advanced data augmentation techniques to cope with data imbalance challenges.

The dataset and fine-tuned models are available for unrestricted access and can be freely downloaded under the Creative Commons Attribution 4.0 International License (CC BY 4.0) from our GitHub repository.\textsuperscript{\ref{ft:github}}

\section{Acknowledgements}\label{sec7}
We sincerely thank Alma Murzagulova, Aizhan Seipanova, Meiramgul Akanova, Almas Aitzhan, Aigerim Boranbayeva, and Assel Kospabayeva, who acted as moderators during the review collection process.
Their tireless efforts, diligence, and remarkable patience contributed significantly to the successful completion of this endeavour.

\section{Bibliographical References}\label{references}

\bibliographystyle{references_style}
\bibliography{references}

\end{document}